\documentclass[sigconf]{acmart}

\usepackage{caption}
\usepackage{subcaption}

\usepackage{amssymb}
\usepackage{dsfont}
\usepackage{multirow}
\usepackage{mathrsfs}
\usepackage{amsmath}
\usepackage{xcolor}
\usepackage{tikz}
\usepackage{bm} 
\usepackage{algorithm}
\usepackage{algorithmic}
\usepackage[dvipsnames]{xcolor}
\usepackage{tikz}
\usepackage[most]{tcolorbox}
\usepackage{multirow}

\newcommand{\na}{\textcolor{blue}} 

\definecolor{namecolor}{HTML}{E6953B}
\definecolor{nametypecolor}{HTML}{4DAF4A}
\definecolor{topologycolor}{HTML}{D65F5F}
\definecolor{topologyhmmcolor}{HTML}{5B2C87}

\definecolor{lightgreen}{rgb}{0.56, 0.93, 0.56}
\AtBeginDocument{%
  }

\setcopyright{acmlicensed}
\copyrightyear{2026}
\acmYear{2026}
\acmDOI{XXXXXXX.XXXXXXX}
\acmConference[arXiv'26]{arXiv}{March 2026}{NY, USA} 

\acmISBN{XXX-X-XXXX-XXXX-X/XXXX/XX}

\begin{document}




\title[A Context Engineering Framework for Improving Enterprise AI Agents based on Digital-Twin MDP]{A Context Engineering Framework for Improving Enterprise \\ AI Agents based on Digital-Twin MDP}%



\author{Xi Yang, Aur\'{e}lie Lozano, Naoki Abe, Bhavya, Saurabh Jha}
\author{Noah Zheutlin, Rohan R. Arora, Yu Deng, Daby M. Sow}

\affiliation{
  \institution{IBM Software Innovation Lab}
  \streetaddress{1101 Kitchawan Road}
  \city{Yorktown Heights}
  \state{New York}
  \postcode{10598}
  \country{USA}
}

\email{{xi.yang, bhavya, saurab.jha, noah.zheutlin, rohan.arora}@ibm.com}
\email{{aclozano, nabe, dengy, sowdaby}@us.ibm.com}



\renewcommand{\shortauthors}{ }

\begin{abstract}

Despite rapid progress in AI agents for enterprise automation and decision-making, their real-world deployment and further performance gains remain constrained by limited data quality and quantity, complex real-world reasoning demands, difficulties with self-play, and the lack of reliable feedback signals. To address these challenges, we propose a lightweight, model-agnostic framework for improving LLM-based enterprise agents via offline reinforcement learning (RL). The proposed Context Engineering via DT-MDP (DT-MDP-CE) framework comprises three key components: (1) A Digital-Twin Markov Decision Process (DT-MDP), which abstracts the agent's reasoning behavior as a finite MDP; (2) A robust contrastive inverse RL, which, armed with the DT-MDP, to efficiently estimate a well-founded reward function and induces policies from mixed-quality offline trajectories; and (3) RL-guided context engineering, which uses the policy obtained from the integrated process of (1) and (2), to improve the agent's decision-making behavior. As a case study, we apply the framework to a representative task in the enterprise-oriented domain of IT automation. Extensive experimental results demonstrate consistent and significant improvements over baseline agents across a wide range of evaluation settings, suggesting that the framework can generalize to other agents sharing similar characteristics in enterprise environments.


\end{abstract}



\maketitle

\section{Introduction}

Large language model (LLM)-based AI agents have emerged as powerful tools for complex reasoning tasks, achieving notable success in well-defined domains such as mathematical reasoning, code generation, and game playing \cite{huang2024understanding,mohammadi2025evaluation}. In enterprise-oriented settings, such as Information Technology operations (ITOps) \cite{mulongo2024key}, however, their effectiveness has been more limited, owing to four distinctive characteristics, including limited quality and quantity of data for model training, the complexity of real-world reasoning tasks, the difficulty of self-play, and the scarcity of verifiable feedback signals. The first two, in particular, pose a fundamental dilemma: training a model for complex tasks generally demands large volumes of data, yet enterprise domains often provide only limited, noisy, or proprietary datasets. These constraints are further amplified by the nature of enterprise environments, where processes frequently involve humans in the loop, making automated data collection difficult, and where sensitive operational data cannot be freely used for training.

Recent advances in LLMs, such as the strong reasoning performance of models like DeepSeek \cite{guo2025deepseek}, have been driven in part by large-scale fine-tuning approaches, including supervised fine-tuning \cite{dong2024abilities} and reinforcement learning (RL)-based fine-tuning \cite{shao2024deepseekmath}. However, the characteristics of enterprise applications limit the effectiveness of these methods. Fine-tuning approaches typically require large amounts of high-quality data, while enterprise tasks are often complex, multi-turn, and dependent on dynamic system states, making data acquisition difficult. In addition, operational constraints often prevent self-play or large-scale online interaction, further restricting RL-based training. Moreover, these methods commonly rely on manually annotated data or task-specific reward designs, which are costly to construct and prone to mis-specification. Consequently, obtaining reliable, verifiable rewards, both at the overall task level and at each step of a multi-turn reasoning process, remains challenging in enterprise settings.

To address these challenges, we propose a comprehensive framework, Context Engineering via DT-MDP \textbf{(DT-MDP-CE)}, for improving LLM-based reasoning agents, built on three key components. \textit{{\bf (1) Digital-Twin Markov Decision Process (DT-MDP)}}: 
LLM agents operate in a Partially Observable MDP (POMDP) with potentially infinite action and state spaces, making direct optimization difficult under limited data. We therefore introduce DT-MDP as an abstract model of the agent, in which its reasoning behavior is represented as a finite MDP, constructed from a modest batch of trajectory data. 
This abstraction can incorporate domain knowledge to capture the most salient aspects of the LLM agent's reasoning behavior, enabling more effective optimization for the task at hand. 
{\textit{{\bf (2) Robust Inverse Reinforcement Learning (Robust-IRL)}}}: 
Abstracting the agent's reasoning process as a finite MDP enables the efficient application of offline inverse reinforcement learning to automatically estimate a well-founded reward function without manual construction. Specifically, we employ a robust \emph{Contrastive Inverse RL} approach, which learns a reward function from pairwise comparisons between trajectories. This allows us to maximally utilize data of limited quantity and quality by leveraging both good and bad ones, thereby {\em improving} the agent rather than merely mimicking them. 
{\textit{{\bf (3) RL-guided Context Engineering}}}: 
The estimated policy learned through the integrated components of (1) and (2) is then used to guide context engineering, enabling targeted interventions that improve the LLM-agent's decision-making behavior.

\begin{figure}
    \centering
    \includegraphics[width=0.48\textwidth]{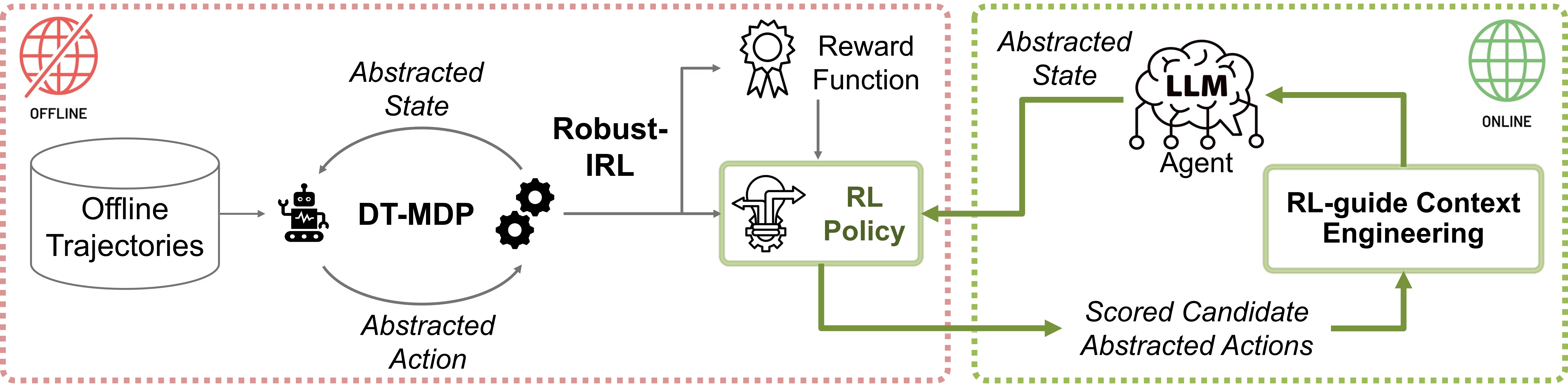}
    \caption{Overall workflow of DT-MDP-CE: an RL policy learned offline through the Digital-Twin MDP (DT-MDP) and robust Contrastive Inverse RL, and applied to the LLM agent via Context Engineering during online execution. 
 }
    \label{figure:overview} 
\end{figure}

The overall workflow of the proposed DT-MDP-CE framework is depicted in Fig. \ref{figure:overview}. It integrates the three components described above: the DT-MDP abstraction is key to the effective application of Robust IRL, which in turn supports the subsequent context engineering. As a case study, we apply the proposed framework on a representative task of Site Reliability Engineering (SRE) in the enterprise-oriented domain of IT automation. Extensive experimental results demonstrate consistent and significant improvements over baseline AI agents across a wide range of evaluation settings, suggesting that DT-MDP-CE can generalize to other AI agents sharing similar characteristics in enterprise environments.

\section{Problem Formulation and Methodology}

\subsection{LLM-based AI Agent}
\label{ref:formulation}


We aim to develop AI agents that perform tasks on behalf of, or in assistance to, various personas in enterprise settings, such as Site Reliability Engineers (SRE) in Information Technology Operations (ITOps). These tasks typically involve multi-turn reasoning, where human operators make sequential decisions while interacting with the environment, for example, by collecting information and executing actions. In the SRE setting, such tasks include diagnosing system failures, identifying root causes of incidents, and implementing remediation strategies. An LLM-based agent designed for this setting therefore performs a multi-step reasoning process, producing a sequence of decisions conditioned on intermediate observations. 

An LLM can be viewed as a function that maps input text to output text. 
When viewed as an agent, the input can be interpreted as \emph{observations}, and the output corresponds to the agent’s internal reasoning step, or \emph{thought}\footnote{This thought might very well be a textual description of an action. The point is that, at this point, it is a potentially infinite text, not a member of a finite set.}. In a sequential decision problem, the LLM-based agent iterates this process multiple times, resulting in a multi-turn reasoning process. 
In its most general form, such an LLM-based reasoning agent can be formalized as implementing a policy $\pi$, that is, a function mapping the entire history of past observations and thoughts into the current output thought: 
\[
\pi: \bar{\Omega} \times \bar{\Theta} \rightarrow \Theta
\]
where we use $\Omega$ and $\Theta$ to denote the set of observations and thoughts, respectively, and  
$\bar{\Omega}$ and $\bar{\Theta}$ their {\em cumulative} counterparts. 

In reasoning tasks such as those in ITOps, observations can be viewed as information about the underlying \textit{environment}, and thoughts correspond to textual descriptions of actions to be taken within that environment. More specifically, observations may reflect some form of \textit{state} of the environment, although the true state may be partially hidden and not readily determined by the observation. The effects of an action, as prescribed by the thought, therefore depend on its underlying state.  
Under this view, we formalize LLM-based agents as operating with a \emph{Partially Observable Markov Decision Process (POMDP)}. 
Consider a standard MDP defined by states $S$ (of the environment), actions $A$ (on to the environment), state transition function $T: S \times A \rightarrow S$, and a reward function $R: S \rightarrow \mathcal{R}$. In the corresponding POMDP, the agent's observations $\Omega$ are determined by a (possibly stochastic) function of states: $\omega: S \rightarrow \Omega$, and the thoughts $\Theta$ are determined by a (possibly stochastic) function of actions: $\theta: A \rightarrow \Theta$. 
This formalization suggests a strategy to {\em estimate} the hidden state and action from the observations and thoughts (the posterior state distribution is often referred to as the \textit{belief state}) and to maximize the expected reward in the underlying MDP with respect to the Bayes posterior distribution \cite{kaelbling98}. Doing so, however, poses a considerable hurdle both in terms of computational and statistical complexity, especially in situations where only limited training data are available. 


The foregoing discussion motivates our proposed approach, namely one in which the hidden state estimation is realized as a deterministic abstraction that maps observations to a finite state space and thoughts to a finite action space, rather than estimating them as posterior distributions. This allows us to have an MDP at hand, where states and actions are visible and finite, capturing salient aspects of the LLM-based agent.  We refer to this resulting MDP as the abstract \emph{Digital-Twin Markov Decision Process (DT-MDP)}. Sequential decision optimization is then performed within this abstract space, and the resulting policy is used to guide and influence the behavior of the underlying LLM-based agent on the target task.


\subsection{Digital-Twin MDP (DT-MDP)}
\label{sec:dt-mdp}

As introduced in Section~\ref{ref:formulation}, we model the multi-turn reasoning process of the LLM-based agent as a \textit{DT-MDP} with finite state and action spaces. The primary motivation is to improve sample efficiency and enable effective learning from limited trajectory data, compared with approaches that operate directly on raw LLM inputs and outputs, which are potentially infinite. Additional benefits of the DT-MDP abstraction include reduced computational complexity for policy optimization and improved interpretability of the agent’s decision rationale. We acknowledge that constructing the DT-MDP requires approximation and heuristics informed by domain knowledge, reflecting a fundamental trade-off between full generality and practical feasibility under data-scarce settings. 


The abstract DT-MDP is constructed from trajectory data generated by the LLM agent it is intended to model. Before abstraction, each trajectory consists of a time-indexed sequence of observations and thoughts produced during a multi-turn reasoning process, together with associated reward signals, such as overall success or failure and, in some cases, intermediate feedback. These raw trajectories are then \emph{abstracted} by mapping observations to abstract observations (which serve as abstract states) and thoughts to abstract actions using a pair of deterministic abstraction functions:
\[
\omega^{-1}: \bar{\Omega} \times \bar{\Theta} \rightarrow O (\approx S)
\]
\[
\theta^{-1}: \Theta \rightarrow A 
\]
where $O$ denotes the space of abstract observations and $A$ denotes the space of abstract actions. The function $\omega^{-1}$ maps cumulative observations and thoughts to abstract observations, and $\theta^{-1}$ maps thoughts to abstract actions. 
Under this abstraction, the original trajectory
\[
\{\langle \omega_t, \theta_t, r_t \rangle\}_{t=1}^T
\]
is converted into an abstract trajectory:
\[
\{\langle s_t, a_t, r_t \rangle\}_{t=1}^T,
\]
where the abstract states $s_t$ and actions $a_t$ are obtained through $\omega^{-1}$ and $\theta^{-1}$, respectively. 
An alternative is to consider the output of $\omega^{-1}$ as an abstract \emph{observation} $o_t$, and treat it as an intermediate step in the process of state abstraction: 
In the latter case, we can consider the whole process of generating the sequences of {\em abstract} observations and actions as a Hidden Markov Model (HMM) and estimate the \textit{hidden} states given the sequence data. 
Specifically, the dynamics defined by the (sampling) policy $\pi$ and the underlying POMDP environment can be formulated as an HMM that generates sequences of the form:\footnote{This type of formulation has been introduced in past work that employed HMM as part of a Q-learning procedure for POMDP \cite{yoon2018}.}
\[
\{ \langle o_t, a_t, r_t \rangle\}_{t=1}^T
\]
with a (hidden) state transition probability function and observation probability function: 
\[
P(s_{t+1} | s_t)
\]
\[
P(\langle o_t, a_t, r_t \rangle | s_t)
\] 
As the reward $r_t$ is assumed to be determined by a (stochastic) function of the observation and action, we simplify the formulation by excluding rewards from the HMM estimation. The resulting HMM model can then be used to estimate the hidden states in any (new) sequence data by a standard decoding procedure such as the Viterbi decoder \cite{viterbi67}.

The resulting abstract trajectory data are sequences of states, actions, and potentially proxies for immediate (or intermediate) rewards for the LLM agent as it performs a task,  
and provide a compact, finite representation of the agent’s behavior, suitable for the subsequent Robust-IRL modeling detailed in Section~\ref{sec:irl}.


\subsection{Robust Inverse Reinforcement Learning}
\label{sec:irl}

Reward design is central to RL for LLMs because it determines what reasoning behaviors the model will acquire. Verifiable rewards (e.g., correctness checks or unit tests) are reliable and scalable but tend to be sparse and delayed. Process Reward Models (PRMs) provide fine-grained, step-level verification that is well-suited to complex reasoning tasks. However, PRMs require dense, step-level supervision, which is expensive and difficult to obtain~\cite{liu2025enhancing}.

The DT-MDP described in Section \ref{sec:dt-mdp} not only addresses the issue of limited data size in our target domain(s), but also the challenge of obtaining verifiable rewards in the trajectories of LLM-based agents. Specifically, the MDP formulation with finite state and action spaces enables the efficient application of the Inverse RL (IRL) approach to derive more objective and robust reward signals. IRL offers a principled way to recover the supervision required by PRMs implicitly from observed behavior. Specifically, IRL aims to determine reward functions from trajectory data without requiring manually designed reward functions~\cite{osa2018algorithmic,arora2021survey}. Its advantages include a reduction of manual labor in reward specification and robustness against mis-specification. In the formulation of DT-MDP, the goal of an IRL method is to refine the per-step rewards $r_t$ associated with the trajectory data.

In its original formulation, IRL aims to estimate a reward function that best explains the behavior observed in \emph{expert demonstrations}, and therefore typically assumes that the input trajectories are optimal~\cite{abbeel2004apprenticeship,ziebart2008maximum,ng2000algorithms,zhang2025understanding}. In our setting, however, the goal is to \emph{improve} the performance of an existing LLM agent rather than simply mimic past behavior. Consequently, we do not assume that the available trajectories are optimal, regardless of their quality (e.g., whether they lead to successful outcomes). Moreover, the number of successful trajectories in enterprise domains is often limited, making it important to leverage all available data for traditional IRL. These considerations motivate the use of a more robust IRL formulation that can learn effectively from mixed-quality trajectories.

\subsubsection{Reward Learning by Contrastive IRL}\label{sec:contIRL}

In view of the above desiderata, we adopt the Trajectory-ranked Reward EXtrapolation (T-REX) \cite{brown2019extrapolating}. T-REX makes use of noisy qualitative trajectory rankings (such as pairwise preferences over demonstrations or ratings for each demonstration on a scale)  to learn a reward neural network $\hat{r}_\theta(s,a)$ that yields higher cumulative returns for higher-ranked trajectories. Here, $\theta$ denotes the neural network's parameters, and $(s,a)$ represents state-action pairs. While traditional IRL methods seek to justify demonstrations, T-REX aims to find a reward function that explains the ranking among them. T-REX can therefore leverage both expert and non-expert trajectories, which is critical when dealing with data of limited quantity and quality. In this paper, we refer to such approaches as \emph{Contrastive IRL}.

The T-REX approach boils down to training a classifier that can predict whether one trajectory is preferable to another based on the predicted returns of each trajectory, following the classic models of preferences ~\cite{bradley1952rank} that have been widely used to train neural networks~\cite{christiano2017deep,rafailov2023direct}. Specifically, given the ranked demonstrations, T-REX seeks to approximate the reward at state $s$ using a neural network, $\hat{r}_\theta(s,a)$, such that $\sum_{s,a \in \tau_i} \hat{r}_\theta(s,a) < \sum_{s,a \in \tau_j} \hat{r}_\theta (s,a)$ when $\tau_i \prec \tau_j$, where $\prec$ indicates the preference between pairs of trajectories. 
The parameters governing the reward function $\hat{r}_\theta$ are estimated using the following loss function, which corresponds to the cross entropy loss applied to a softmax-normalized distribution:{\small
\begin{align*}
 \mathcal{L}(\theta) = -\sum_{\tau_i \prec \tau_j} \log \frac{\exp \left(\sum_{(s,a) \in \tau_j} \hat{r}_\theta(s,a)\right)}{\exp \left(\sum_{(s,a) \in \tau_i} \hat{r}_\theta(s,a)\right) + \exp \left(\sum_{(s,a) \in \tau_j} \hat{r}_\theta(s,a)\right)}.
\end{align*}}
Note that a discount factor can be naturally incorporated into the trajectory returns if desired. The neural network used to approximate the reward function can be chosen flexibly, allowing it to handle discrete, continuous, or high-dimensional state-action representations. In many enterprise LLM-agent settings, qualitative trajectory rankings can be obtained, for example, through LLM-as-a-judge scores that evaluate outcomes or the overall quality of reasoning. Importantly, T-REX relies only on pairwise rankings induced by these scores, rather than on the scores themselves. Therefore, the scores need not be highly precise, and T-REX has been shown to be robust to noise in the resulting rankings~\cite{brown2019extrapolating}. 


\subsubsection{Policy Induction and Selection} \label{sec:IRLpi}
Equipped with a reward function $\hat{r}(s,a)$ (e.g., learned via Contrastive IRL), we estimate the optimal policy $\pi(s,a)$ through offline reinforcement learning using approaches such as Deep Q-Network~\cite{mnih2013playing} or Conservative Q-Learning~\cite{kumar2020conservative}. These methods take as input a set of training trajectories \[\{\{\langle s^i_t,a^i_t,r^i_t\rangle\}_{t=1}^{T_i},i=1,\ldots N_{\textrm{train}}\},\] where $r^i_t$ is the reward for taking action $a^i_t$ at state $s^i_t$, $N_{\textrm{train}}$ is the number of trajectories in the training set, and $T_i$ is the length of the $i^\textrm{th}$ trajectory. For the rewards $r_i^t$, we consider several options, including per-turn rewards estimated by IRL (i.e., $r^i_t=\hat{r}(s^i_t,a^i_t)$), purely outcome-based rewards such as LLM-as-a-judge scores, or combinations of IRL-derived per-turn rewards and outcome scores.


DT-MDP can operate with different state-action abstractions and support various IRL and RL methods with different hyperparameter settings, yielding a potentially large set of candidate policies. Offline evaluation is therefore desirable to prioritize which policies to deploy online, especially in settings where online testing is costly, risky, or time-consuming. For this purpose, we employ Off-Policy Evaluation (OPE)~\cite{uehara2022review}, which estimates the value of each candidate policy using historical data not seen during training, without interacting with the environment. This enables policy selection based on estimated performance before deployment. Specifically, in the DT-MDP setting, we use Fitted Q-Evaluation~\cite{fu2021benchmarks}, which re-estimates the Q-value function of each candidate policy from data collected under a reference policy. This produces less biased value estimates and allows fairer comparisons across policies. The resulting Q-function is then used to compute a summary metric, such as the initial-value score~\cite{paine2020hyperparameter}, which approximates the expected return from initial states. Candidate policies are then ranked according to this score, and the top-$k$ are selected for online evaluation.



\subsection{RL Policy-based Context Engineering}
\label{sec:context_engineering}



We use the RL policy learned from offline abstracted trajectories via DT-MDP to guide the agent’s behavior during online execution. Specifically, at each agent's turn, we obtain an abstract state $s = \omega^{-1}(\bar{\omega}, \bar{\theta})$ from the accumulated interactions between the agent and the environment. The abstract RL policy $\pi$ can be applied to assess the adequacy of all candidate abstract actions $a$ by $\pi(a|s)$. We intervene in the agent’s behavior using context engineering strategies tailored to the specific agent (for example, inserting selected actions $a$ into the prompt), in order to approximately implement the prescribed abstract action in the real environment as best as possible. Concrete examples are provided in Section \ref{sec:caseStudy}.



\section{A Case Study -- SRE Diagnosis Agent} 
\label{sec:caseStudy}



As a case study, we apply the proposed approach to a key task in the enterprise-oriented domain of IT operations automation, namely \emph{Site Reliability Engineering (SRE)}. SRE focuses on detecting software anomalies and facilitating the diagnosis and resolution of enterprise incidents. In this work, we concentrate on the SRE diagnosis task and apply our framework to improve the performance of two LLM-based agents. The first is the domain-specific \emph{Explanations over Graphs (EoG)} agent, a state-of-the-art system designed for this task~\cite{jha2026think}. The second is the general-purpose \emph{ReAct} agent~\cite{yao2022react}. The EoG agent leverages system topology to guide planning during root-cause investigation. At each step, it selects a suspicious entity, collects evidence, and then evaluates whether the entity is the root cause and summarizes the investigation so far. The agent maintains both a per-trajectory queue of candidate entities for future exploration and a per-turn queue for selecting entities to examine at the current step. Achieving accurate and efficient diagnosis requires optimizing the sequence of entities explored during the investigation, making this task well-suited to RL-based intervention.

\subsection{Digital Twin MDP (DT-MDP) for SRE}


\subsubsection{Trajectory Data Processing}

To train the RL policy, we use the agent's trajectories of multi-turn interactions with the environment while diagnosing an incident. We process the raw trajectory data to extract relevant information and transform them into abstract (state and action) representations, as required by the DT-MDP formulation. Specifically, from the software development kit (SDK) logs of historical trajectories \cite{crouse2023formally}, we extract turn-level state and action information by prompting an LLM. The abstract state is derived from the diagnosis findings summarized by the agent up to the current turn, while the abstract action corresponds to the entity selected for exploration at that turn. In addition, since inverse RL methods require quality signals over trajectories, we apply an LLM-as-a-judge to assess their quality \cite{jha2025itbench}. The judge compares the agent’s outputs against the ground-truth fault propagation to evaluate aspects such as the identified root cause and fault conditions.


\subsubsection{State and Action Representations}
\label{dt_mdp_example}

Parsing and processing the agent’s trajectory data, as described above, provides the information needed to construct the state and action representations of the DT-MDP. The specific design of these representations is a modeling choice that should be guided by domain knowledge. For the present SRE case study, we experiment with several DT-MDP abstraction variants, described in detail below.


\indent $\bullet$ \textit{Name-based:} In this scheme, we use the name of the entity to be explored at each step as the abstract action, represented using 1-hot encoding. The state is expressed with a similar vector over entities, where each entry reflects the agent’s current assessment of that entity as a \textit{primary} failure, \textit{cascading} failure, or \textit{normal}. These categories are encoded using the values 2, 1, and 0, respectively.

\indent $\bullet$ \textit{Name-type-based:} It is analogous to the name-based scheme above. Instead of using vectors whose elements correspond to the entity names, vector elements now represent entity name-type pairs, resulting in a considerably higher dimensionality. As before, 1-hot encoding is used for the action representation, while the state representation uses multi-valued coding with values 2, 1, and 0.

One thing to note about name-based and name-type-based representation schemes is that the entity names (and their types) under consideration are explicitly encoded in the representations. As a result, the learned model does not extend to applications having distinct sets of entities. To the extent possible, it would be desirable to \emph{relativize} the state and action representations so that the learned model can be applied to scenarios where different or new entities are involved, motivating our third representation scheme, which builds on, extends, and generalizes the first two schemes.


\indent $\bullet$ \textit{Topology-based:} Recognizing that the topological structure among entities is often crucial for SRE diagnosis task, we develop a multi-dimensional feature representation scheme based on the underlying topology graph. In this approach, each entity is represented by features that characterize the structure of its topological neighborhood, particularly in relation to entities currently judged to be in failure and those where initial anomalies were observed. 

We consider the following {\em distance}-based state and action features. Here, the \emph{distance} between two entities is defined as the length of the shortest path between their corresponding nodes in a directed graph, where an edge represents the presence of event propagation. For \textit{action features}, we use the distance between the current and previous entities; the distance between the current entity and the \emph{symptom} entity (i.e., the first entity examined after the initial alert); the distance between the next target entity and entities identified as primary failures; and the distance between the next target entity and entities identified as cascading failures. For \textit{state features}, we use the distance between the symptom entity and entities identified as primary failures, and the distance between the symptom entity and entities identified as cascading failures.
We also optionally consider one other action feature, which is more global in nature, the \emph{Hubs} score for the target entity of the action \cite{kleinberg1999}. 

In addition to using the above representations directly as the state representations, we have also treated them as intermediate abstract observations and applied HMM modeling to estimate the hidden abstract states using the topology-based state and action feature representations as the HMM output observations. 
In our experiments, we primarily consider the distance-based ones as the \textit{Topology-based} features. 
The Hubs feature, as well as the HMM feature, are used only in those experiments investigating the effects of feature enrichment, as reported in Section~\ref{sec:feature}.

\subsection{Robust Inverse RL and RL for SRE} 

To estimate the rewards for subsequent RL modeling, we employed Contrastive IRL as described in Section~\ref{sec:contIRL}. For trajectory rankings, we consider two signals, including (1) the \emph{Fault Propagation Chain Accuracy} score: an LLM-as-a-judge metric ranging from 0 to 100, which evaluates the agent's ability to identify the causal path, namely the path from the symptom to the entity identified as the root cause. Specifically it is the F1 score for the accuracy of the identified fault propagation chain; (2) the average between the \emph{Root Cause Entity Identification} score and the \emph{Fault Propagation Chain Accuracy} score, where the former is an LLM-as-a-judge, `all-or-nothing' score in $\{0,100\}$ that checks if the response generated by the agent correctly identifies the primary root cause entity. In our experiments, we found that it is valuable to take into account not only the success/failure of a trajectory but also the \textit {logic} of the trajectory as reflected by the \emph{Fault Propagation Chain Accuracy}. Given a set of LLM-as-a-judge scores, optimizing for the most informative combination is a pertinent direction for future work.



To induce the policies and associated long-term reward functions, we considered Conservative Q-learning (CQL) ~\cite{kumar2020conservative} and Behavior Cloning~\cite{syed2012imitation,raza2012teaching,gleave2022imitation}, using the implementations provided in~\cite{d3rlpy}. We also experimented with alternative approaches, such as Soft Actor Critic~\cite{haarnoja2018soft}, but the training was less stable. Overall, this lead to three groups of policies: \emph{RL-IRL}, which applies CQL using per-turn rewards predicted by contrastive IRL; \emph{RL-Sparse}, which applies CQL using Root Cause Entity Identification as a sparse final reward; and \emph{BC}, which uses behavior cloning.



\subsection{RL-based Context Engineering for SRE} 
\label{sec:ce_sre}

Our context engineering strategies are illustrated using two SRE agents: the EoG agent \cite{jha2026think} and the ReAct agent \cite{yao2022react}. 

\begin{figure}
    \centering
    \includegraphics[width=0.79\linewidth]{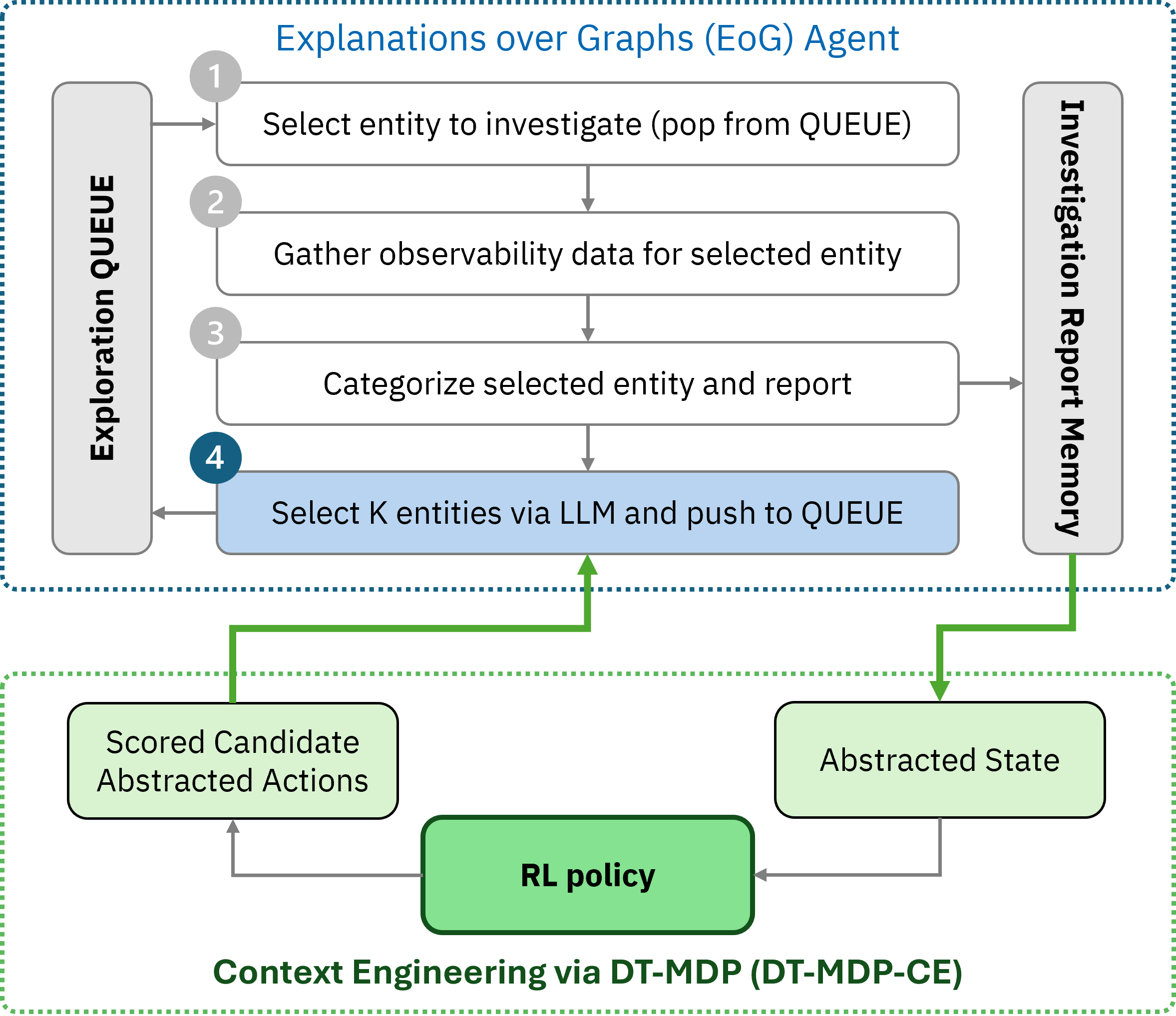}
    \caption{Illustration for EoG Agent with DT-MDP-CE.}
    \label{figure:EoG}
\end{figure}

\subsubsection{EoG Agent} Explanations over Graphs (EoG) agent explores a set of candidate entities connected by an underlying topology, with the goal of identifying the root cause of an incident \cite{jha2026think}. Given this set-up, we equate the set of entities to the abstract action space. At each agent turn, we derive an abstract state $s = \omega^{-1}(\bar{\omega}, \bar{\theta})$ and then apply the abstract RL policy $\pi$ to compute the action probabilities $\pi(a|s)$. 
With the EoG agent, we intervene the agent’s decision-making at the fourth step (highlighted in blue in Fig. \ref{figure:EoG}) via one or more of the following context engineering strategies.

\indent $\bullet$ \textit{I: Suggesting via Prompts.} Given the RL policy's score (or selection probability) for each of the candidate actions, we include those entities having probabilities exceeding a certain threshold as suggestions in the prompt when the LLM agent is asked to select the next suspicious entities to be added to the exploration queue (Step 4 of the EoG flow). An example prompt is shown as follows: 

\begin{tcolorbox}[
  colback=gray!5,
  colframe=black!60,
  arc=2mm,
  boxrule=0.8pt
]
\textbf{Example of \textit{Suggesting via Prompts}:}

... \textit{The actions related to \{RL suggested entities\} are often relevant in this scenario, so lean toward exploring them if they show up in the above observed evidence} ...
\end{tcolorbox}



\indent $\bullet$ \textit{II: Pruning Explorations.} After the LLM agent identifies the most suspicious entities, and before adding these candidates to the exploration queue (step 4 of the EoG flow), we assess the necessity of exploring each entity using the probabilities predicted by the RL policy. Entities whose probabilities fall below a predefined threshold are filtered out and not added to the queue.

\indent $\bullet$ \textit{III: Prioritizing Explorations.} Before adding the suspicious entities identified by the agent to the exploration queue, we reorder them based on the probabilities predicted by the RL policy. Entities with higher probabilities are assigned higher priorities and placed earlier in the queue, so they are explored first in subsequent turns.

\subsubsection{ReAct Agent} Reasoning and Acting (ReAct) is a widely used, general-purpose agentic framework that interleaves reasoning and action by generating intermediate thoughts to guide decision-making while interacting with external tools through a Thought-Action-Observation loop \cite{yao2022react}. Though the ReAct agent differs from EoG in its underlying mechanisms, our DT-MDP formulation can be readily applied to abstract states and actions from the agent’s observations and thought-action pairs to learn the RL policy. Consequently, the three RL-based context engineering strategies described for EoG can also be adopted to ReAct. For strategy \textit{I: Suggesting via Prompts}, entities whose probabilities exceed a predefined threshold are added as suggestions to the prompt before the agent generates the next thought–action pair. For strategy \textit{II: Pruning Explorations}, after the ReAct agent produces a thought-action pair, the probability of the corresponding DT-MDP-abstract action is evaluated using the RL policy; if it falls below the threshold, the action is skipped. For strategy \textit{III: Prioritizing Explorations}, at each turn, the agent generates multiple candidate thought–action pairs; the RL policy evaluates them and executes the candidate with the highest probability.


\section{Experiments}
\label{sec:results}

To evaluate the effectiveness of the DT-MDP-CE framework, we assess its performance on the SRE diagnosis task \cite{jha2025itbench} using the EoG agent as the primary sandbox. To further examine generalization, we extend the evaluation to an alternative ReAct-based SRE diagnosis agent, an additional enterprise software engineering (SWE) task, and LLMs from both the Mistral and Llama families at different scales. Additionally, we analyze the stability and robustness of the framework under diverse parameter settings.

\subsection{Datasets} 

The data leveraged in our experiments are collected from a benchmark with IT applications (ITBench) \cite{jha2025itbench}, which aims to evaluate AI agents for IT automation use cases with interpretable metrics. Our training set consists of agent-system interaction trajectories collected from 12 SRE diagnosis scenarios in ITBench: four \textit{Flagd} scenarios involving feature-flag-induced failures, seven \textit{Chaos Mesh} scenarios generated via fault injection (e.g., pod failures and network disruptions), and one \textit{Customized} IT failure scenario representing a domain-specific enterprise incident. Each scenario falls into one of these categories. Using LLMs of varying sizes as SRE diagnostic agents (including GPT-4o, GPT-OSS-120B, GPT-OSS-20B, O4-Mini, Mistral-Medium-2505, and Granite-3.2-8B-Instruct), we collected 819 trajectories (12,079 turns), each representing a multi-turn diagnostic process conducted by the agent. 


\subsection{Online Evaluation}
\label{online-eval}


The agent is evaluated online on six ITBench test scenarios, including one Flagd failure, one Chaos Mesh failure, and four customized failures unseen during offline training. For each scenario, we repeat the diagnosis process 15 times. Each trial is evaluated using the LLM-as-a-judge protocol from ITBench \cite{jha2025itbench}, where Gemini-2.5-Pro \cite{comanici2025gemini} compares predicted results against ground-truth root causes. Performance is measured using Pass@3 recall and F1, which count a scenario as successful if at least one of three trials produces a correct diagnosis. To ensure fair comparisons, we fix the agent’s underlying LLM to Mistral-Medium-2505. The RL policies used for context engineering are trained over offline trajectories collected from the training scenarios. Based on the off-policy evaluation procedure described in Section~\ref{sec:IRLpi}, we select the RL-IRL policy learned with Conservative Q-Learning (CQL) \cite{kumar2020conservative}, using per-turn rewards predicted by contrastive inverse reinforcement learning, where trajectory rankings are derived from root-cause and fault-propagation correctness. This selected policy ranks among the top three candidate policies in 87.5\% of the offline evaluation settings across different RL models and reward combinations.



\subsubsection{Digital Twin MDP (DT-MDP)} 
\label{online-eval-dtmdp}
\begin{figure}
    \centering
    \includegraphics[width=1.0\linewidth]{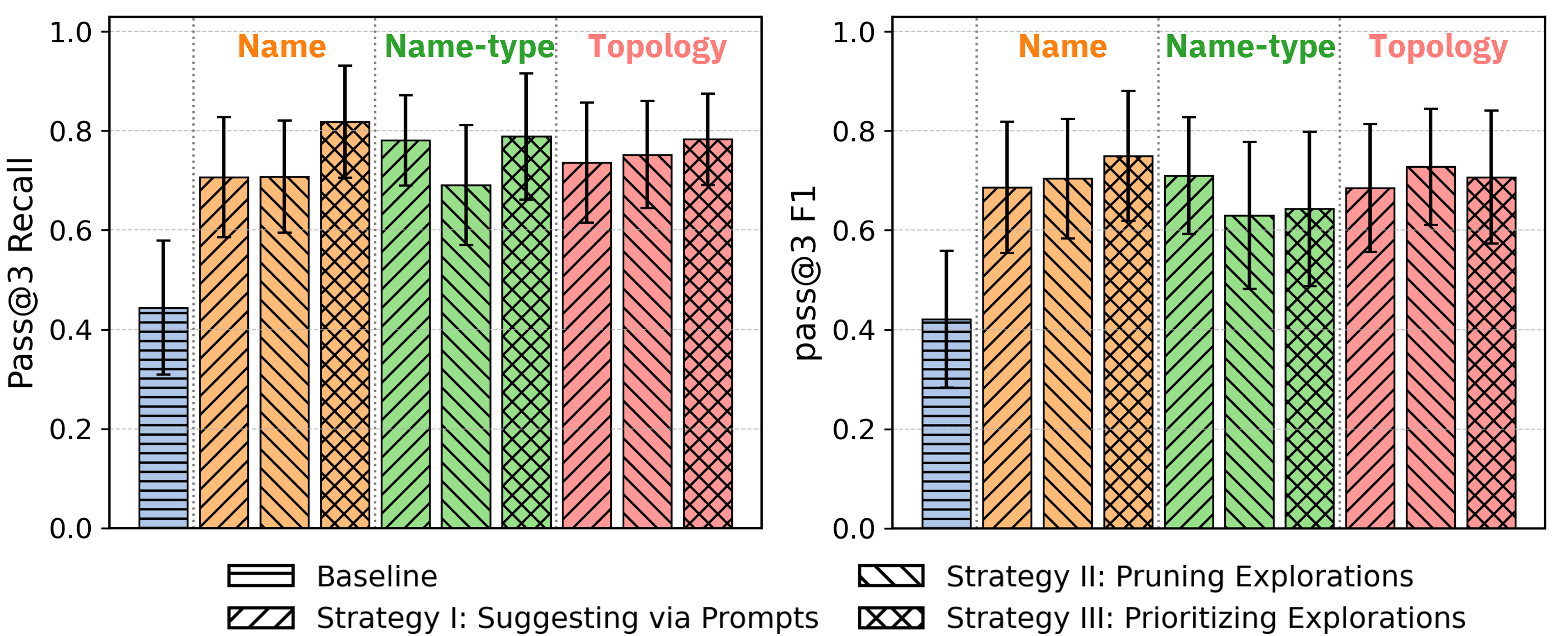}
    \caption{Comparing EoG agent with various RL-based CE strategies learned from \textcolor{namecolor}{\textbf{Name}}-, \textcolor{nametypecolor}{\textbf{Name-type}}-, and \textcolor{topologycolor}{\textbf{Topology}}-based DT-MDP vs. baseline (original agent without CE).}
    \label{fig:compared_baseline}
\end{figure}

Fig.~\ref{fig:compared_baseline} presents the performance of Context Engineering via DT-MDP (DT-MDP-CE) with three DT-MDP variants (Name-, Name-type-, and Topology-based) under three context engineering strategies, compared against the baseline EoG agent without RL-based CE. To extract the DT-MDP representation, we prompt Mistral-Medium-2505 to perform entity extraction, and then convert the extracted information into vectorized representations as described in Section~\ref{dt_mdp_example}. Results are reported using Pass@3 Recall (left) and Pass@3 F1 (right). The values are estimated via a bootstrap Monte Carlo procedure: for each test scenario, three trials are randomly sampled with replacement, and the process is repeated 200 times to obtain stable estimates. Across both metrics, DT-MDP-CE consistently outperforms the baseline, demonstrating the effectiveness of CE guided by RL policies learned with DT-MDP. We further perform Bonferroni-corrected paired t-tests \cite{bonferroni1936teoria} against the baseline across the test scenarios. All Name-type and Topology-based configurations show statistically significant improvements after correction (p<0.05), while the Name-based configurations do not reach significance despite showing numerical gains. Overall, the performance differences among the Name-type and Topology-based variants and across strategies remain small. This likely reflects the substantial semantic overlap encoded by the DT-MDP variants, which allows the EoG agent to infer missing topological structure and reduces the impact of explicit representation differences. Additionally, the strategies address similar exploration bottlenecks, leading to partially overlapping effects.

\begin{table}[t]
\centering
\small
\setlength{\tabcolsep}{3pt}   
\renewcommand{\arraystretch}{0.95} 
\caption{Input \& output tokens and diagnosis time for EoG agent with DT-MDP variants using CE strategies (I/II/III).}
\begin{tabular}{l|c|ccc|ccc|ccc}
\toprule
 & \multirow{2}{*}{\textbf{Baseline}}
 & \multicolumn{3}{c|}{\textbf{Name}}
 & \multicolumn{3}{c|}{\textbf{Name-Type}}
 & \multicolumn{3}{c}{\textbf{Topology}} \\
\cmidrule(lr){3-5}
\cmidrule(lr){6-8}
\cmidrule(lr){9-11}
\textbf{Cost}
& 
& \textbf{I} & \textbf{II} & \textbf{III}
& \textbf{I} & \textbf{II} & \textbf{III}
& \textbf{I} & \textbf{II} & \textbf{III} \\
\midrule
In Tok.(K)
& 439 
& 525 & 412 & 554 
& 601 & 451 & 539 
& 571 & 305 & 541 \\

Out Tok.(K)
& 4.7 
& 4.6 & 3.1 & 5.7 
& 4.8 & 3.0 & 6.1 
& 4.9 & 3.2 & 5.1 \\

Time (s)
& 841 
& 886 & 818 & 924
& 968 & 869 & 888 
& 986 & 778 & 894 \\
\bottomrule
\end{tabular}

\label{tab:cost}
\end{table}

\begin{figure}
    \centering
    \includegraphics[width=0.9\linewidth]{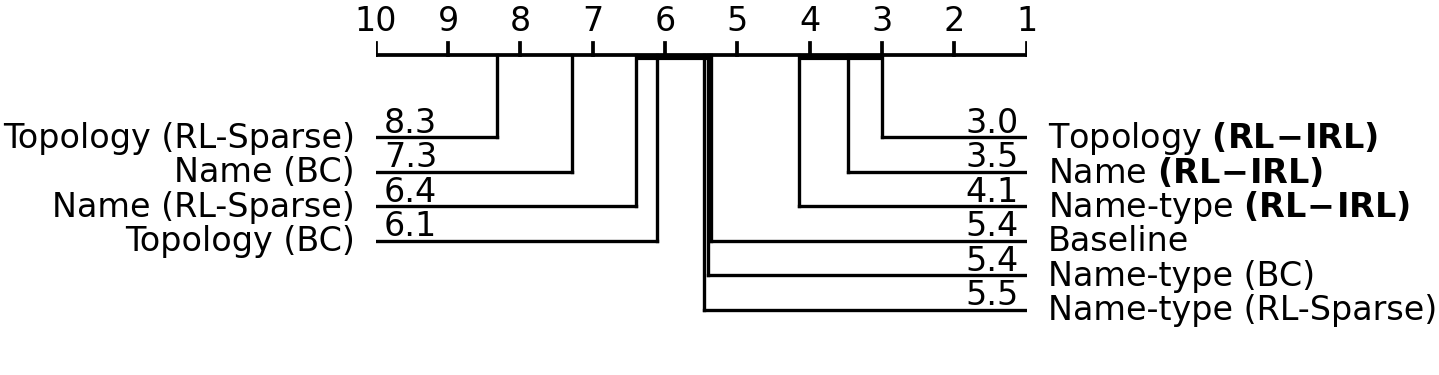}
    \caption{CD Analysis: CE based on RL-IRL, RL-Sparse, and BC Policies vs. Baseline (original EoG without CE).}
    \label{fig:CDGraph}
\end{figure}

\subsubsection{Robust Inverse Reinforcement Learning} We conduct a Critical Difference (CD) analysis using the Nemenyi test \cite{demvsar2006statistical} to compare the average ranks of different methods across test scenarios. In the CD diagram, methods are ranked per scenario, and those whose average ranks differ by less than the critical difference are considered statistically indistinguishable. Lower ranks indicate better performance, and methods not connected by horizontal lines are significantly different. We compare CE with RL policies learned from IRL-derived rewards (\textit{RL-IRL}), RL with sparse final rewards (\textit{RL-Sparse}), Behavior Cloning (\textit{BC}), and the \textit{Baseline} agent without CE across DT-MDP variants (Name-, Name-type-, and Topology-based), as shown in Fig.~\ref{fig:CDGraph}. The diagram shows that the RL-IRL group consistently achieves the best average ranks, while RL-Sparse, BC, and baseline methods tend to occupy lower ranks and are often clustered together, indicating similar performance. This suggests that IRL-derived intermediate rewards provide more informative learning signals, leading to more effective policy induction.


\begin{figure}
    \centering
    \includegraphics[width=1.0\linewidth]{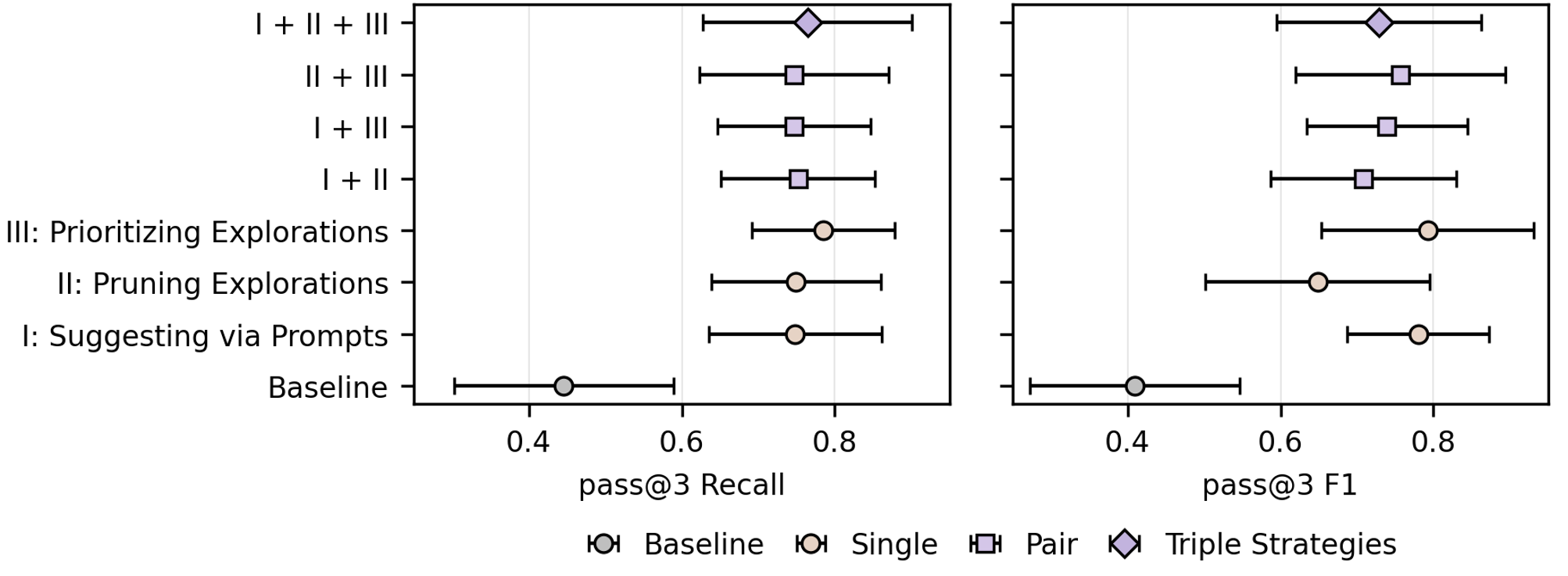}
    \caption{Comparing three Context Engineering strategies and their pairwise \& triple combinations for EoG agent.}
    \label{fig:combinations}
\end{figure}

\subsubsection{RL-based Context Engineering} To evaluate the effectiveness of each CE strategy, we test them individually as well as in pairwise and triple combinations under Topology-based representations. The Pass@3 recall and F1 scores, estimated via bootstrap Monte Carlo sampling, are shown in Fig.~\ref{fig:combinations}. The baseline without RL performs substantially worse than all RL-enhanced settings. Among the single strategies, Strategy III achieves the best average performance. Combining strategies does not consistently improve performance: pairwise and triple combinations are often comparable to single-strategy results. This may be because all three strategies intervene in the agent’s behavior in relatively mild and similar ways. As a result, their individual effects are comparable, and combining them does not yield substantial additional gains. We also evaluate the cost (i.e., input/output tokens and diagnosis time) in Tab. \ref{tab:cost}. Overall, our DT-MDP-CE introduces only modest overhead compared to the baseline, and Strategy II can even reduce the cost by pruning unnecessary explorations.

\subsection{Generalization}


\begin{figure}
    \centering
    \includegraphics[width=1.0 \linewidth]{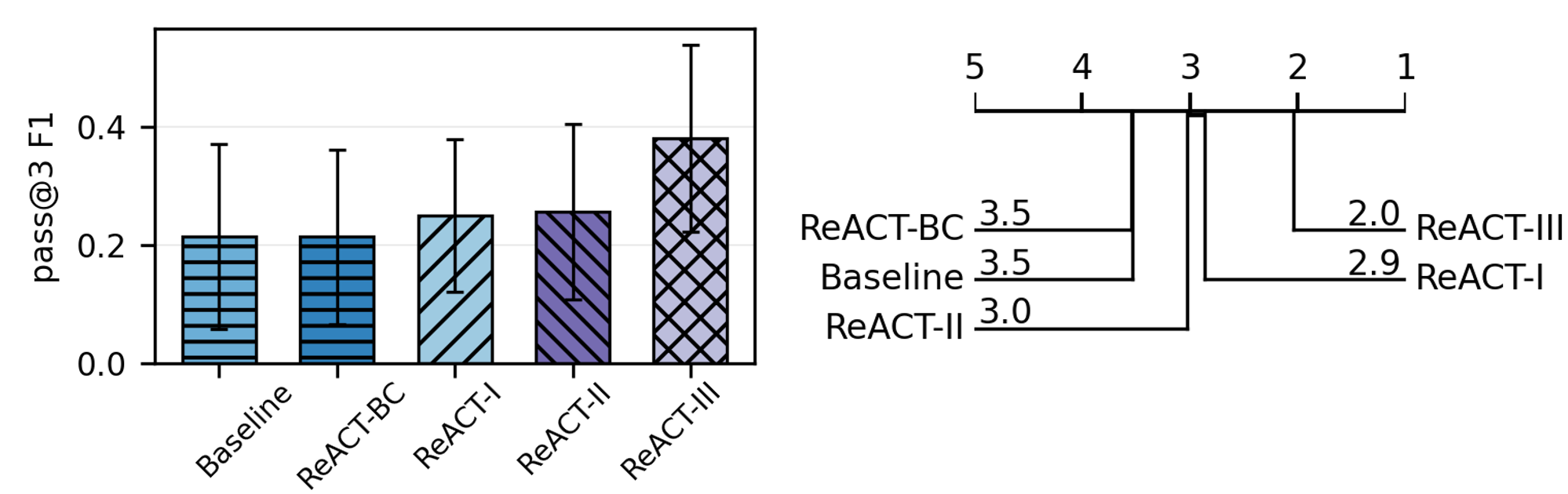}
    \caption{ReAct with three RL-based CE strategies (I–III) vs. Behavior Cloning (BC) and baseline (original ReAct agent).}
    \label{fig:react_results}
\end{figure}

\subsubsection{Agents} In addition to the EoG agent, we apply our DT-MDP-CE to a ReAct agent \cite{yao2022react} for the SRE diagnosis task. We compare three strategy variants (ReAct-I, ReAct-II, and ReAct-III) against two baselines: ReAct with behavior cloning (ReAct-BC using Strategy I) and the original ReAct agent, using online evaluation. Fig.~\ref{fig:react_results} shows the Pass@3 scores (left) and the CD analysis (right). RL-based context engineering consistently improves performance over the baselines, with ReAct-III achieving the best Pass@3 F1 score. Strategy III performs slightly better in ReAct, possibly because its design implicitly incorporates aspects of the other two strategies by selecting the most recommended action from the RL policy while pruning the remaining candidates.




\subsubsection{Domains} 
\begin{figure}
    \centering
    \includegraphics[width=1.0\linewidth]{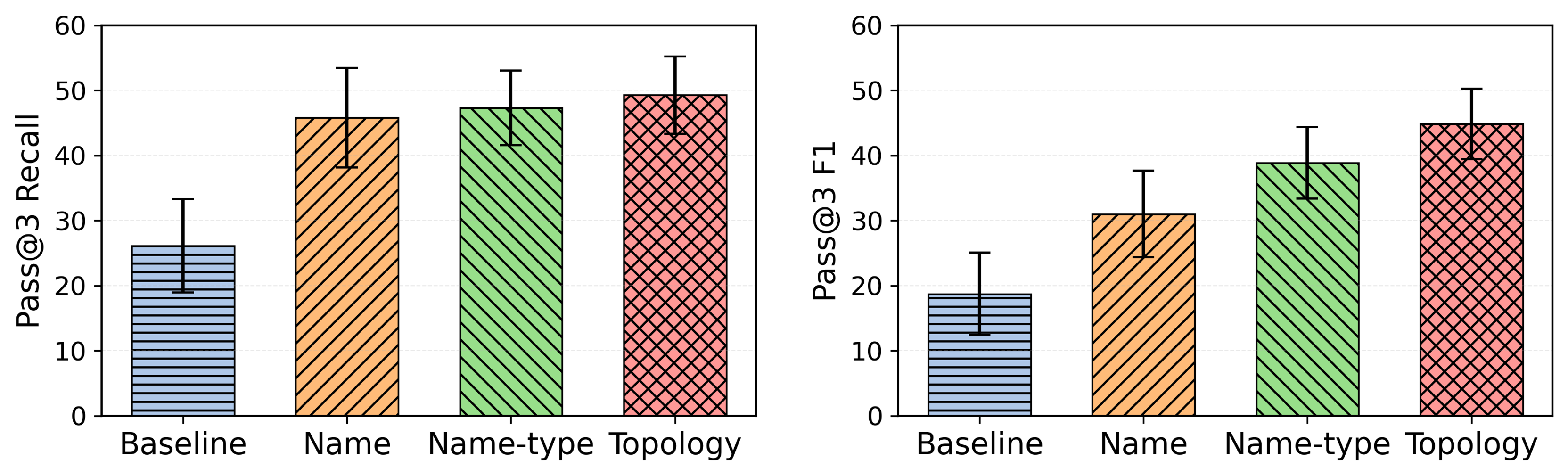}
    \caption{Comparing agent with CE based on DT-MDP variants vs. Baseline without CE over SWE scenarios. }
    \label{fig:code}
\end{figure}

To assess generalization beyond SRE diagnosis, we evaluate the framework in the Software Engineering (SWE) domain, where scenarios come from the same application but involve different, code-related failure mechanisms. 
We apply RL policies learned from the same SRE incidents and test them on these code-based scenarios to assess how the learned policies transfer without re-training. We compare Pass@3 recall and F1 on for DT-MDP variants as shown in Fig.~\ref{fig:code}. Across both metrics, all DT-MDP-based variants outperform the baseline without RL, indicating effective transfer to the SWE setting. Performance improves from the Name-based to the Name-type and Topology-based abstractions, with the Topology variant achieving the best average results. This trend suggests that richer and relativized structural representations help the transferability of the DT-MDP learned policies. 

\subsubsection{LLM Models}

\begin{figure}
    \centering
    \includegraphics[width=1.0\linewidth]{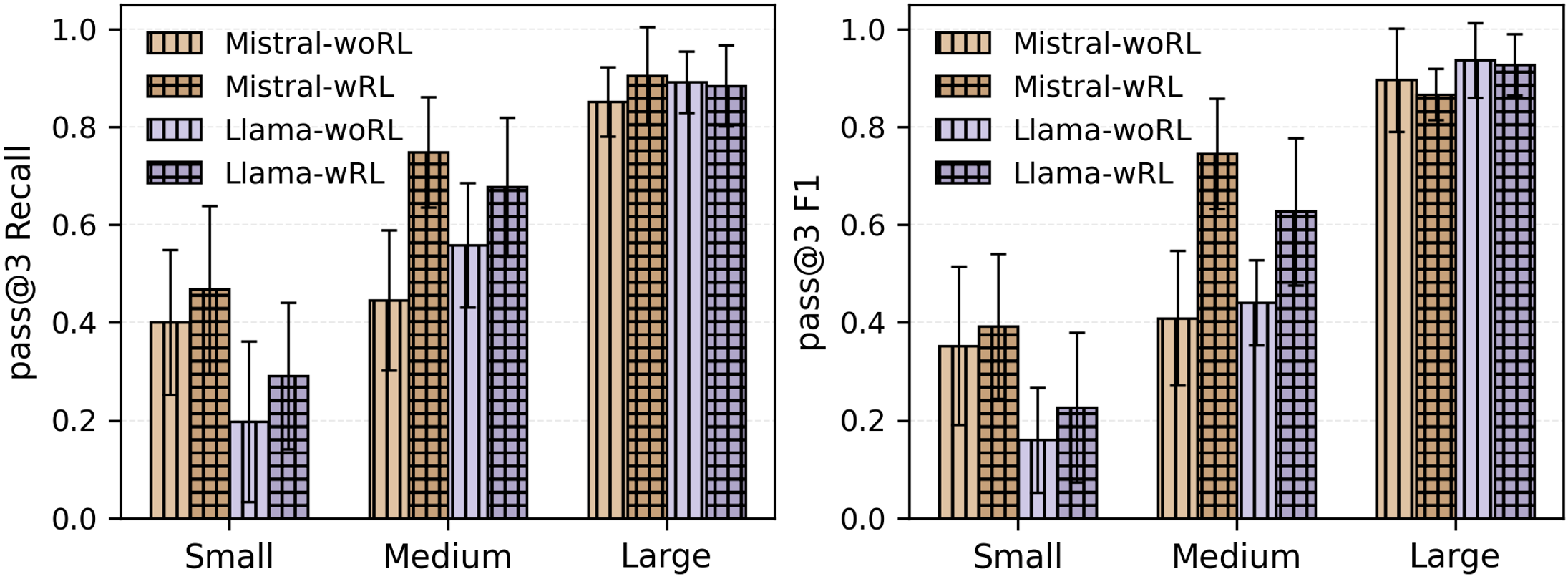}
    \caption{Comparing with vs. without RL (wRL vs. woRL) with different model sizes across Mistral and Llama families.}
    \label{fig:model_size}
\end{figure}

We further analyze the impact of RL-based context engineering across different model scales and families by comparing performance with and without it (denoted ``wRL" and ``woRL") for \textit{Small}, \textit{Medium}, and \textit{Large} models from the Mistral and Llama series. Specifically, we evaluate Mistral-Small-3.1-24B-Instruct-2503, Mistral-Medium-2505, and Mistral-Large-Instruct-2407 (123B), as well as Llama-3.1-8B, Llama-3.3-70B-Instruct, and Llama-3.1-405B-Instruct-FP8. Fig.~\ref{fig:model_size} shows how the effectiveness of context engineering varies with model capacity. Across both families, RL-based context engineering consistently improves performance, with the largest gains at the medium scale. Small models benefit modestly, likely due to limited capacity to exploit the interventions. Medium-sized models show the strongest improvements, suggesting a favorable balance between model capacity and RL guidance. In contrast, gains for large models are smaller, as they already achieve strong baseline performance without RL.

\subsubsection{Feature Enrichment} 
\label{sec:feature}

\begin{figure}
    \centering
    \includegraphics[width=1.0\linewidth]{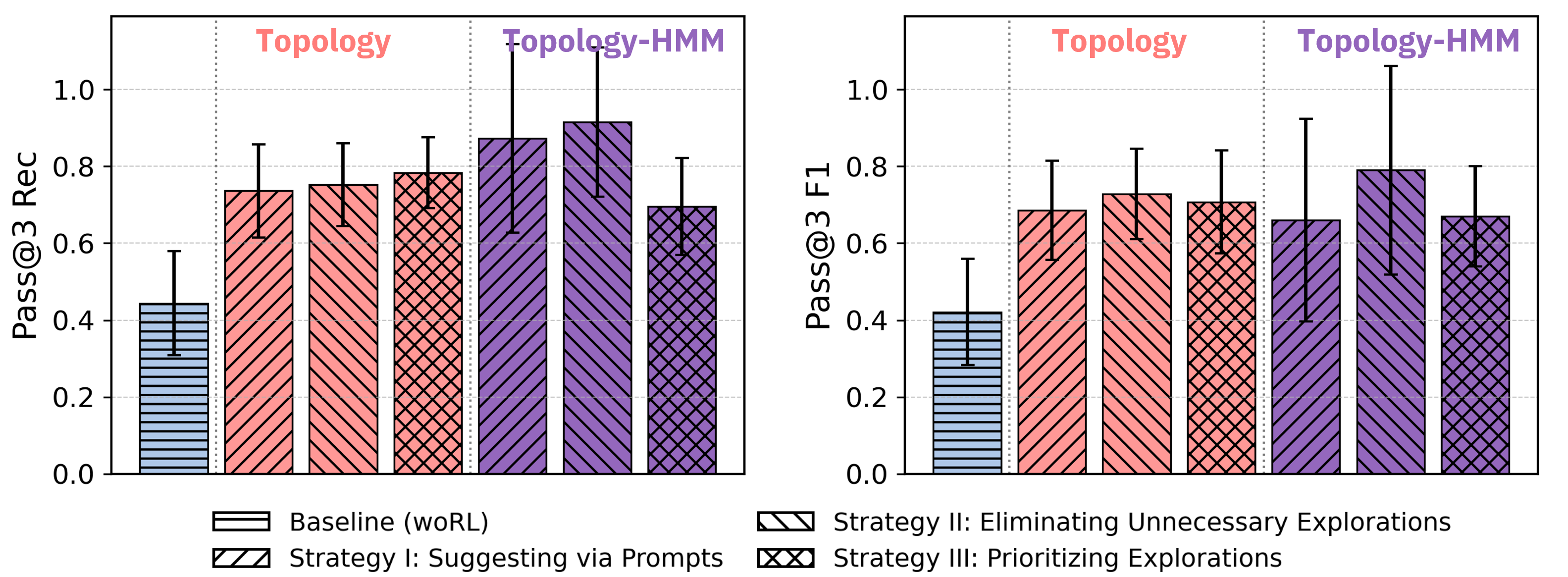}
    \caption{Comparing EoG agent with CE strategies learned from original \textcolor{topologycolor}{\textbf{Topology}}- and \textcolor{topologyhmmcolor}{\textbf{Topology-HubHMM}}-based DT-MDP vs.\ baseline (original EoG agent without CE).}
    \label{fig:hubhmm}
\end{figure}

As mentioned in Section \ref{dt_mdp_example}, the features comprising the state and action representations in DT-MDP are a design choice that is domain dependent and can be subjective. To investigate the sensitivity and robustness of this design choice, we further experimented with an enhanced version of the Topology-based representations for actions and states.
Specifically, we first augment the distance-based features with the \emph{Hubs} feature reflecting the more global topological properties (c.f.\ Section~\ref{dt_mdp_example}). We then treat the resulting Topology-based features as intermediate abstract observations and apply HMM modeling to estimate the hidden abstract states. The resulting HMM states are treated as an additional feature. The results are presented in Figure \ref{fig:hubhmm}. The progressively richer feature representations (with distances, Hubs and HMM states) appear to contribute to better accuracy overall, and over the baseline with significant margins, (albeit having larger variances.) It is reasonable to postulate that richer and carefully constructed structural models can lead to more effective search and decision strategies, while the advantage of DT-MDP over the baseline is relatively robust with respect to minor variations in its representational design choice.

\subsection{Key Parameters}

\subsubsection{Number of Expert Trajectories in Robust IRL} 

\begin{figure}
    \centering
    \includegraphics[width=1.0 \linewidth]{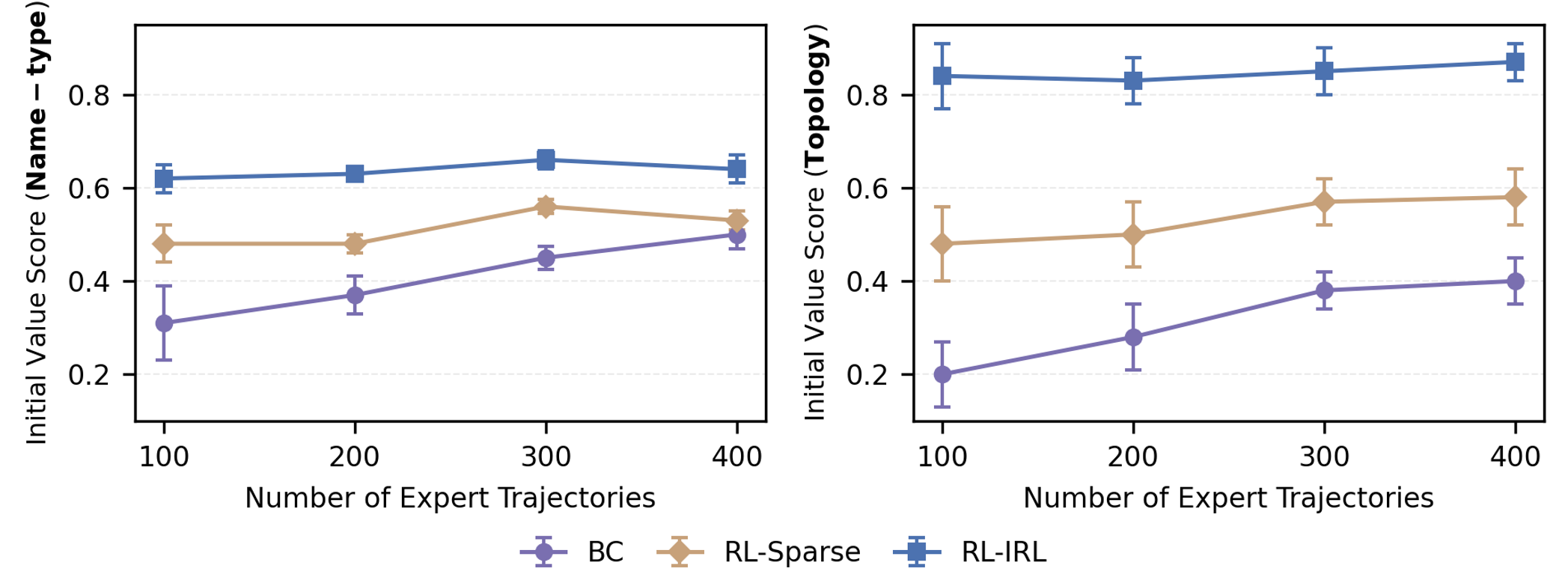}
    \caption{Robustness of Name-type- and Topology-based policy learning with increased number of expert trajectories.}
    \label{fig:traj_num}
\end{figure}

We evaluate robustness with respect to the number of expert trajectories (i.e., successful diagnoses) using off-policy evaluation \cite{uehara2022review}. We randomly sample \{100, 200, 300, 400\} successful trajectories for training and report the initial-value score, which approximates the expected reward from initial states. The higher the initial-value score, the better the policy. Fig.~\ref{fig:traj_num} shows results for Name-type (left) and Topology-based (right) policies learned with CQL using IRL-derived rewards (\textit{RL-IRL}), sparse rewards (\textit{RL-Sparse}), and Behavior Cloning (\textit{BC}). BC is highly sensitive to the number of trajectories, while RL, especially RL-IRL, remains more robust and consistently achieves the best performance, aligning with the results observed in Fig.~\ref{fig:CDGraph}. 


\subsubsection{Thresholds in Context Engineering}

\begin{figure}
    \centering
    \includegraphics[width=1.0 \linewidth]{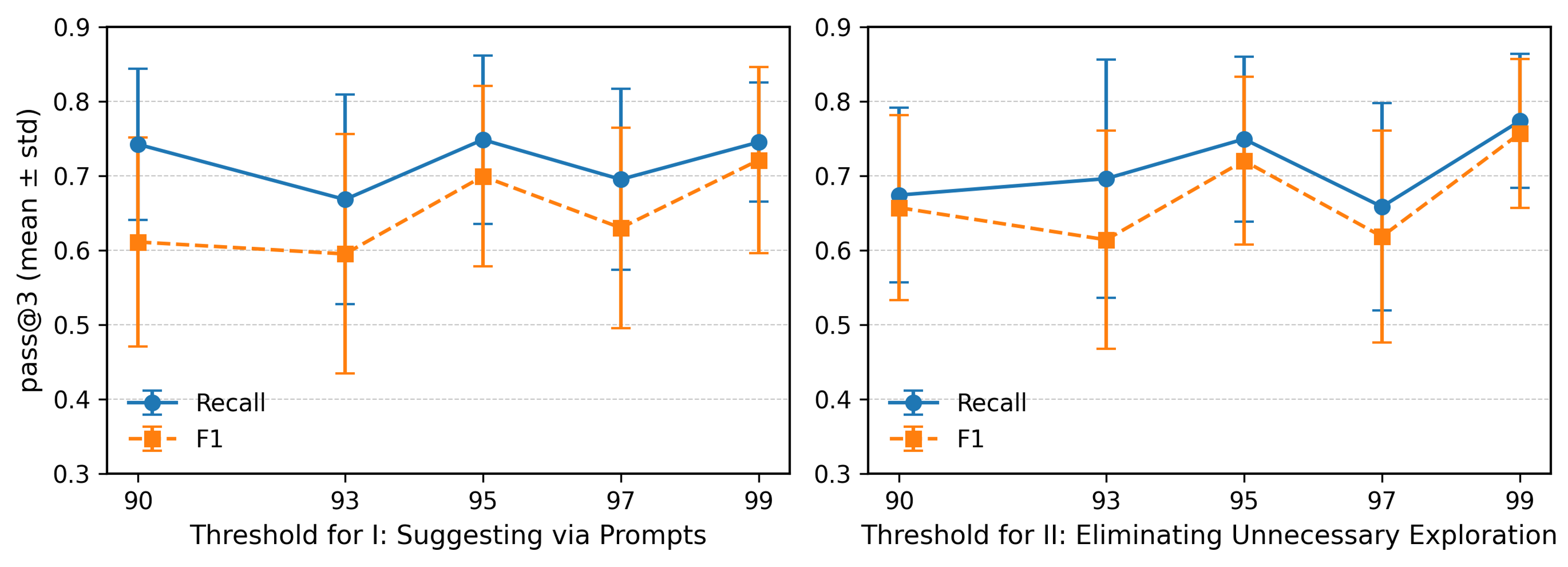}
    \caption{Robustness of thresholds for CE strategies I \& II.}
    \label{fig:thresholds}
\end{figure}

We set the default thresholds in Strategies I and II for suggesting and pruning actions to 95 and 85 percentiles among the candidates, respectively. To assess robustness, Fig.~\ref{fig:thresholds} shows the sensitivity of the framework to these parameters. For each strategy, we vary the threshold while keeping other settings fixed, and report Pass@3 Recall and F1 with std. The results remain stable across the tested ranges for both strategies, with only small variations relative to the observed variance, indicating the framework is generally robust to threshold selection.

\section{Related Works}

\textit{\textbf{Reward Design for LLMs: }} Reward design determines what reasoning behaviors an LLM acquires during RL. Existing approaches include verifiable rewards (e.g., correctness checks), generative or learned rewards, dense process-level rewards, and unsupervised intrinsic signals \cite{zhang2025survey}. Verifiable rewards are reliable but generally sparse, while dense or learned rewards require costly annotations or risk bias. In contrast, our framework estimates rewards directly from mixed-quality trajectories via contrastive inverse RL, reducing reliance on handcrafted or densely annotated reward signals.

\noindent \textit{\textbf{Context Engineering: }} Context engineering studies how to select, structure, and update the information provided to LLMs, including retrieved knowledge, memory, and tool outputs \cite{mei2025survey,wang2024learning,jeong2024adaptive}. Recent methods treat context construction as a sequential decision problem, using RL or bandit algorithms to choose relevant documents, tools, or prompt tokens \cite{zhang2025rearank,deng2022rlprompt,luo2025agent}. Unlike these approaches, which apply RL directly in high-dimensional text spaces, our method first builds a finite DT-MDP abstraction of the agent’s reasoning process, enabling sample-efficient offline learning before applying the resulting policy to guide context engineering.

\noindent \textit{\textbf{Model-based Interventions for LLM Agents: }} Most LLM agents are model-free, relying on contextual reasoning without explicit representations of environment dynamics \cite{yao2022react,schick2023toolformer,trivedi2023interleaving}. Model-based approaches instead incorporate world models or simulators to guide planning and evaluation \cite{m2023model,ferrag2025llm}. Our DT-MDP abstraction serves as a lightweight behavioral model learned from trajectories. Rather than modeling the full environment, it captures key reasoning dynamics in a finite MDP, enabling practical offline policy learning in data-constrained enterprise settings.

\noindent \textit{\textbf{Supervised and RL Fine-tuning: }} Supervised fine-tuning (SFT) and RL-based fine-tuning are the dominant approaches for adapting LLMs to specialized domains \cite{dong2024abilities,shao2024deepseekmath}. SFT is generally stable and effective, but it requires large volumes of high-quality labeled data, which are costly and difficult to obtain in enterprise settings \cite{shen2024rethinking}. RL-based fine-tuning, on the other hand, relies on reliable reward signals and often requires online interaction or self-play to collect training data \cite{zha2025rl}, which may be impractical, unsafe, or restricted in enterprise environments. These limitations make large-scale fine-tuning difficult to deploy in such domains. Our approach avoids extensive fine-tuning and instead improves agent behavior through efficient offline inverse RL and policy-guided context engineering, requiring only limited trajectory data and no online interaction.

\section{Discussions \& Conclusions}
\label{sec:conclusions}

In this paper, we proposed a framework for improving LLM-based agents for enterprise-oriented tasks based on the DT-MDP formulation. Through extensive empirical evaluation in the domain of IT automation, we showed that RL-based context engineering guided by DT-MDP policies consistently improves agent performance. The framework is general and agent-agnostic, as it operates on abstract state-action representations rather than task-specific logic. This design suggests its applicability to a broader range of enterprise scenarios and other settings with similar characteristics. More broadly, DT-MDP-CE can be viewed as a form of model-based context engineering for reasoning LLMs. By incorporating structured models through context engineering, the approach improves the reliability, interpretability, and alignment of LLM agents, offering a pathway toward more controllable and trustworthy systems in complex real-world settings. This integration of linguistic reasoning and structured modeling forms a hybrid architecture that balances flexibility with precision, and points to a promising direction for building LLM agents that are not only capable, but also controllable, explainable, and dependable in practice.


For future work, we plan to apply DT-MDP-CE across a broader range of SRE scenarios, as well as more diverse enterprise applications. The four characteristics we identified in enterprise settings (e.g., limited data, complex reasoning demands, difficulty of self-play, and scarce feedback) also arise in other domains, such as healthcare, where DT-MDP-CE may provide similar benefits. In addition, we plan to extend the DT-MDP framework to fine-tuning. In particular, we aim to leverage the long-term cumulative reward estimates produced by the DT-MDP formulation to guide supervised fine-tuning and further improve agent performance.

\balance
\bibliographystyle{ACM-Reference-Format}
\bibliography{literatures}

\appendix

\newpage

\section{Appendix}



\subsection{Additional Discussion of Related Work}

\noindent \textbf{Reward Design for LLMs. } Reward design is the foundation of RL for LLMs, as it dictates what kind of reasoning behaviors an LLM will learn. Rewards in general can be categorized into verifiable, generative, dense, and unsupervised \cite{zhang2025survey}. Specifically, verifiable rewards, such as correctness checks or unit test outcomes, are the most scalable and reliable, making them ideal for tasks with objectively measurable outcomes, while they are often sparse and delayed, providing limited learning signals.  Generative or learned rewards, which rely on model-based or AI feedback when explicit correctness is unavailable, offer greater flexibility yet risk bias and reward exploitation. Dense and process-oriented rewards deliver feedback on intermediate reasoning steps, helping models learn how to reason rather than only optimizing final answers. Unsupervised rewards, derived from intrinsic signals such as novelty, uncertainty, or self-consistency, are valuable when external supervision is limited. Effective reward shaping, which blends multiple reward signals to balance accuracy, coherence, and diversity, is essential for stabilizing training and enhancing generalization across reasoning tasks. Unlike prior reward design approaches that rely on manually specified signals, verifiable checks, or learned reward models, our method derives rewards automatically from trajectory rankings via contrastive inverse RL. This allows reward estimation from sparse, noisy, and mixed-quality enterprise trajectories without requiring step-level supervision or handcrafted rewards. \\

\noindent \textbf{Inverse RL. } 
As mentioned in Section~\ref{sec:contIRL}, the most intuitive offline apprenticeship learning is behavior cloning (BC) \cite{syed2012imitation,raza2012teaching,gleave2022imitation}, which involves building a mapping from states to actions to imitate demonstrated actions \cite{pomerleau1991efficient}. However, it highly relies on the quality and quantity of the demonstrations \cite{ross2011reduction,shao2024offline}. Instead of merely imitating demonstrated actions, inverse reinforcement learning (IRL) \cite{klein2011batch, mori2011model, jain2019model, lee2019truly, chan2021scalable, oh2019sequential} and generative adversarial imitation learning \cite{kostrikov2018discriminator, kostrikov2019imitation} can induce policies more robustly by either explicitly or implicitly learning the reward function. 
\emph{IRL}-based methods typically involve a computational online iterative loop, including inferring an explicit reward function, inducing a policy using traditional RL, rolling out the learned policy, and updating the reward parameters based on the divergence between the roll-out behaviors and demonstrations. 
The rewards in IRL-based methods are usually modeled with a tractable format, such as a linear function \cite{abbeel2004apprenticeship,ziebart2008maximum,babes2011apprenticeship}, mapping states or state-action pairs to reward values. 
To avoid rolling out the policy, batch-IRL methods have been proposed \cite{raghu2017continuous}. 
\emph{Generative adversarial imitation learning}-based methods (GAIL) usually involve iteratively learning a generator to roll out the policy and a discriminator to distinguish the learned behaviors from the demonstrations. To avoid rolling out the policy in a batch setting, some off-policy learning methods have been developed based on off-policy actor-critic algorithms \cite{kostrikov2018discriminator,kostrikov2019imitation}. However, they inherit the complex alternating max-min optimization from general adversarial imitation learning \cite{ho2016generative}. 
The quality of demonstrations matters for inducing accurate policies. In addition to the contrastive IRL we leverage for contrastive IRL(\cite{brown2019extrapolating,brown2020better}),  methods proposed to handle sub-optimal demonstrations include \cite{wu2019imitation,tangkaratt2020variational,wang2021learning,wang2021robust,yang2021trail,xu2022discriminator,wang2023unlabeled,bu2023learning}. Many of these are based on GAIL and/or involve online components. Compared with traditional IRL and imitation-learning methods that assume optimal demonstrations or require online rollouts, our approach employs contrastive IRL within the DT-MDP abstraction to learn from mixed-quality, fully offline trajectories. This design avoids expensive online interactions and enables robust reward estimation in data-scarce enterprise settings. \\

\noindent \textbf{Context Engineering. } Context engineering encompasses the methods used to select, structure, and manage the information provided to large language models, shaping how they reason and perform across diverse tasks. Rather than viewing context as a single, static prompt, context engineering treats it as a dynamic assembly process that integrates retrieved knowledge, memory, tools, intermediate outputs, and task-specific signals into a coherent input \cite{mei2025survey}. This perspective frames context as an optimization target, determining what information to include, how to format it, and when to update or discard it, under practical constraints such as context length, latency, and reliability. By treating context as a core component of system design, context engineering aims to improve grounding, adaptability, and reasoning by optimizing what the model sees, how it sees it, and when it sees it.

Within this broader landscape, planning-based or policy-based context selection methods treat context construction as a sequential decision-making problem optimized through learned policies. Instead of using fixed retrieval heuristics, these approaches employ reinforcement learning, bandit algorithms, or value-based estimators to decide which context elements to fetch, prioritize, or exclude to maximize downstream performance. Examples include RL-based retrieval policies such as Learning to Retrieve (L2R) \cite{wang2024learning}, Adaptive RAG systems \cite{jeong2024adaptive} that learn how many and which documents to include, bandit-based reranking \cite{zhang2025rearank} using Q-learning or UCB to choose evidence chunks, and agent frameworks where models learn which tools or information sources to invoke as the task unfolds \cite{schick2023toolformer}. These decision-theoretic approaches enable adaptive, context-efficient systems that respond to task dynamics rather than relying on static retrieval pipelines. While prior context engineering work focuses on retrieval, tool selection, or heuristic context construction, our framework treats context generation as a policy-driven intervention guided by an offline-learned RL policy. The DT-MDP enables structured, model-based context decisions rather than relying solely on retrieval heuristics or online exploration. \\

\noindent \textbf{Model-based Interventions for LLM Agents. } 
LLM agents have emerged as powerful general-purpose reasoning systems capable of performing diverse tasks such as dialogue, planning, coding, and multi-step problem solving \cite{yao2022react,schick2023toolformer,trivedi2023interleaving}. Their ability to integrate natural language understanding with procedural reasoning has opened new avenues for autonomous agents that can plan and act through linguistic instructions. Despite these advances, current LLM agents remain predominantly model-free: they rely on pattern recognition and contextual inference without an explicit understanding of the environment’s dynamics or the causal structure underlying their actions. This reactive nature limits their capacity for foresight, consistency, and safe decision-making, particularly in dynamic or high-stakes settings \cite{huang2024understanding}.

In contrast, model-based reasoning, a principle long established in fields such as reinforcement learning and control theory, emphasizes the use of explicit world or behavioral models to simulate, predict, and evaluate outcomes before acting \cite{m2023model}. Applying this paradigm to LLM agents introduces the concept of model-based interventions: external or embedded models that can guide, monitor, or modify an agent’s reasoning process. These interventions may take the form of learned world models, formal logic constraints, simulators, or meta-predictive components that evaluate the plausibility, safety, or efficiency of an agent’s plans. By incorporating such structured models, LLM agents gain the ability to reason about ``what-if” scenarios, anticipate the results of their actions, and adjust behavior proactively rather than reactively \cite{ferrag2025llm}. Integrating model-based interventions into LLM agents has the potential to significantly enhance reliability, interpretability, and alignment. This combination of linguistic reasoning and structured modeling creates a hybrid architecture that leverages the strengths of both paradigms and language-based flexibility and model-based precision. Ultimately, model-based interventions offer a promising pathway toward building LLM agents that are not only capable but also controllable, explainable, and trustworthy in complex real-world environments. In this paper, we aim at introducing the interventions to LLM agents taking the power of RL models. Our work instantiates this model-based intervention paradigm through the DT-MDP abstraction, which provides an explicit decision model learned from agent trajectories. Unlike prior approaches that rely on simulators or external world models, our method constructs a compact behavioral model directly from offline data and uses it to guide agent reasoning via context engineering. \\

\noindent \textbf{Supervised and RL Fine-tuning. } 
Although general-purpose LLMs exhibit broad generalization, they often underperform in specialized domains due to domain shift, limited domain-relevant data, and mismatches between pretraining objectives and task-specific requirements. Fine-tuning (FT) enables models to incorporate domain-specific knowledge, constraints, and reasoning patterns that are absent from generic pretraining. In general, fine-tuning approaches can be broadly categorized into two types: supervised fine-tuning (SFT) \cite{dong2024abilities} and reinforcement learning (RL)–based fine-tuning \cite{shao2024deepseekmath}.

SFT is relatively stable and reproducible to implement; however, it typically relies on large volumes of high-quality, human-curated data, which are time-consuming and labor-intensive to collect. In many enterprise scenarios, e.g., AI for IT Operations (AIOps), access to such large-scale, high-quality datasets is limited. Moreover, LLMs are increasingly required to function as agents that perform multi-step tasks involving multiple rounds of tool calling, inference, and generation, which further complicates data curation for supervised fine-tuning, as annotations are often needed at each interaction step. Although models can be trained on multi-turn conversations, existing approaches frequently treat single turns equally despite their varying importance. Additionally, overly long multi-turn inputs can degrade fine-tuning performance by introducing truncation, diluting supervision signals, and increasing noise in optimization, particularly in agent-based settings. In practice, SFT alone usually plateaus before aligning with downstream goals.

RL-based fine-tuning typically requires well-defined reward signals to guide learning and often relies on online interaction or self-play to improve performance through trial and error. Furthermore, executing policies for self-play and reward collection may be infeasible, restricted, or potentially unsafe in certain domains, such as enterprise environments, thereby limiting the practicality of existing RL-based fine-tuning approaches. In contrast to supervised or RL fine-tuning approaches that require large-scale labeled data, reliable rewards, or online interaction, our framework operates entirely offline and does not modify the underlying LLM parameters. Instead, it improves agent behavior through inverse RL and policy-guided context engineering, making it suitable for data-limited and safety-critical enterprise environments. \\

\subsection{Prompts for DT-MDP}

\subsubsection{EoG Agent} 

The abstract action (i.e., focal entity) is parsed from the SDK logs \cite{crouse2023formally}, and the abstract actions are extracted by LLM (Mistral-Medium-2505) with the following prompt: 

\begin{tcolorbox}[
  colback=gray!5,
  colframe=black!60,
  arc=2mm,
  boxrule=0.6pt,
  left=3pt,right=3pt,top=3pt,bottom=3pt
]
\small
\textbf{Task:} Extract structured (name, type) entities from the incident description and classify them as:

\textbf{Primary Failure} (root cause), 
\textbf{Cascading Failure} (secondary impact), or 
\textbf{Normal} (healthy).

Each entity must be a tuple such as \texttt{("frontend","Pod")}.

\textbf{Rules:}
Only use names from \texttt{\{filtered\_entities\}} and types from \texttt{\{filtered\_types\}}.  
Use only plausible name–type pairs.  
A name must appear verbatim in the sentence.  
Do not infer or hallucinate names or types.  
Each pair appears in only one category.

If no entities are present, return:
\begin{verbatim}
{"primary_failure":[],"cascading_failure":[],"normal":[]}
\end{verbatim}

\textbf{Output:}
Return only valid JSON with keys
\texttt{"primary\_failure"}, \texttt{"cascading\_failure"}, and \texttt{"normal"}.

\textbf{Sentence:} \texttt{\{query\}}
\end{tcolorbox}

\subsubsection{ReACT Agent} 

We extract the abstract state with the following prompt:

\begin{tcolorbox}[
  colback=gray!5,
  colframe=black!60,
  arc=2mm,
  boxrule=0.6pt,
  left=3pt,right=3pt,top=3pt,bottom=3pt,
  breakable
]
\small
\textbf{Task:} You are an expert in microservices and Kubernetes (\texttt{kubectl}).
Determine the state of each observability modality (\textit{metrics}, \textit{logs}, \textit{traces}) for
\texttt{\{entity\_name\_type\}}, using \textbf{only} the Tool Output produced by \texttt{\{tool\_name\}}.

\textbf{Rules:} Use only the Tool Output. Do not infer or assume. No explanation. Output JSON only.
If a modality is not explicitly mentioned, set it to \texttt{"None"}.

\textbf{Output (exact JSON):}
\begin{verbatim}
{
  "metrics": "Normal" | "Abnormal" | "None",
  "logs": "Normal" | "Abnormal" | "None",
  "traces": "Normal" | "Abnormal" | "None"
}
\end{verbatim}

\textbf{Tool Output:} \texttt{\{tool\_output\}}
\end{tcolorbox}

The abstract actions are extracted with the following prompt: 

\begin{tcolorbox}[
  colback=gray!5,
  colframe=black!60,
  arc=2mm,
  boxrule=0.6pt,
  left=3pt,right=3pt,top=3pt,bottom=3pt,
  breakable
]
\small
\textbf{Task:} You are an expert in microservices and Kubernetes (\texttt{kubectl}).
Extract exactly one \texttt{(name, type)} entity pair that the agent is about to investigate,
using both the tool input and the agent’s reasoning.

\textbf{Allowed Entities:}  
Names (must appear verbatim, case-sensitive): \texttt{\{filtered\_entities\}}  
Types (assign only from this list): \texttt{\{filtered\_types\}}

\textbf{Extraction Rules:}
\begin{itemize}
\item A name is valid only if it appears verbatim in the agent thought or tool input.
\item Do not infer or hallucinate names.
\item After identifying a valid name, choose the most appropriate type from the allowed list.
\item Be conservative; do not invent or assign all types by default.
\item The type need not appear in the text; infer only from context.
\item Prioritize: (1) Tool input, (2) Agent thought.
\end{itemize}

\textbf{Output:}
\begin{verbatim}
("name", "type")
\end{verbatim}
No markdown or explanations.  
If multiple names appear, choose the one most central to the tool input.  
If none appear, output:
\begin{verbatim}
("NONE", "NONE")
\end{verbatim}

\textbf{Text:}  
Agent thought: \texttt{\{agent\_thought\}}  
Tool input: \texttt{\{agent\_input\}}
\end{tcolorbox}


\subsection{Additional Details on the Experiments}

\noindent \textbf{Contrastive IRL and RL hyperparameters.}
Following~\cite{brown2019extrapolating}, we trained neural networks for contrastive IRL with three fully connected layers with ReLU nonlinearities. For the Name and Name-type abstractions, we used 256 hidden units, while for the Topology representation we additionally experimented with 16 hidden units due to the smaller feature dimensionality. Similarly, for CQL training we used the default networks with 256 hidden units for the Name and Name-type configurations, while we also experimented with 16 hidden units for the Topology-based representation.

\end{document}